\documentclass{IEEEtran}
\usepackage{setspace}
\usepackage{lipsum}

\usepackage{filecontents}
\usepackage{filecontents,lipsum}
\usepackage[noadjust]{cite}
\usepackage{amsmath}
\usepackage{nccmath}
\usepackage{stfloats}
\usepackage{array}
\usepackage{float}
\usepackage{subfig}
\usepackage{amsmath}
\usepackage{multirow}

\usepackage{amssymb}
\DeclareMathOperator*{\argmin}{arg\,min}
\DeclareMathOperator*{\argmax}{arg\,max}

\usepackage{cite}
\usepackage{amsmath,amssymb,amsfonts}
\usepackage{algorithmic}
\usepackage{graphicx}
\usepackage{textcomp}
\begin{document}

\title{Unsupervised Segmentation of Fire and Smoke from Infra-Red Videos}
\author{
Meenu Ajith and
Manel Mart\'{i}nez-Ram\'on, \IEEEmembership{Senior Member, IEEE}
\thanks{Department of Electrical and Computer Engineering, The University of New Mexico, New Mexico 87106, USA (e-mail: majith@unm.edu)}
\thanks{This work has been supported by NSF S\&CC EAGER grant 1637092.}
\thanks{Corresponding author: Meenu Ajith (e-mail:majith@unm.edu).}}

\maketitle

\begin{abstract}
This paper proposes a vision-based fire and smoke segmentation system which use spatial, temporal and motion information to extract the desired regions from the video frames. The fusion of information is done using multiple features such as optical flow, divergence and intensity values. These features extracted from the images are used to segment the pixels into different classes in an unsupervised way. A comparative analysis is done by using multiple clustering algorithms for segmentation. Here the Markov Random Field performs more accurately than other segmentation algorithms since it characterizes the spatial interactions of pixels using a finite number of parameters. It builds a probabilistic image model that selects the most likely labeling using the maximum a posteriori (MAP) estimation. This unsupervised approach is tested on various images and achieves a frame-wise fire detection rate of 95.39\%. Hence this method can be used for early detection of fire in real-time and it can be incorporated into an indoor or outdoor surveillance system.
\end{abstract}


\section{Introduction}
The fire and smoke detectors are an important part of firefighting systems and are also widely used in monitoring indoor buildings and outside environments. The conventional detection systems use inbuilt sensors which do not issue the alarm unless the particles reach the sensors to activate them. To obtain a high precision, the sensors must be distributed densely in close proximities. Hence in real-world situations, these are highly inefficient, and the delayed response may cost the life of firefighters and other human lives. As an appropriate alternative to conventional methods, vision-based fire and smoke detection systems were introduced in the past few years.\par
The vision-based systems either utilize the color information of fire and smoke or it uses the dynamic motion features \cite{1,2}. The classical approaches operated on RGB \cite{3}, YCbCr \cite{4}, CIE L*a*b \cite{5} or HSI \cite{6} color spaces and the image pixels were classified according to the appearance model of the fire.  But the use of color information gives high false alarm rates due to similar colors present in the surrounding environments. Another approach involving disorder analysis and growth rate was used to minimize these false alarms \cite{7,8}. Further, one of the most popular methods was wavelet analysis, which used various extracted features for the detection process. The wavelet domain energy analysis was done to analyze the time variant behavior of the smoke \cite{9}. To improve the performance, an Expectation Maximization based GMM model was obtained by training the pixels of previously occurred events\cite{10}. In \cite{11}, on the other hand, used a change detection algorithm by extracting foreground pixels. Nevertheless, the system suffers heavily from a change in illumination and hence require fine-tuning of the algorithm parameters. In \cite{12}, the optical flow vectors were calculated based on the turbulence characteristics of the smoke and were used to eliminate non-smoke disturbances. Other methods calculated the motion direction of smoke by assuming grayscale invariance in the optical flow algorithm \cite{13}. Recent papers also modeled a Markovian process by considering the motion of the fire \cite{14}. Later, Hidden Markov models \cite{15}  were used to distinguish between the flame and flame-colored objects. These models evaluated the spatial color variations in flame to reach a final decision.\par
The state-of-the-art architecture of a fire/smoke detector can be summarized in three steps such as pixel-wise classification of fire/smoke, region-based segmentation and the analysis of these regions. In this paper, a comparative analysis of different segmentation algorithms is done to find the appropriate one for fire and smoke detection. The experiments are conducted on Infrared (IR) datasets available online. Moreover, different feature extraction methods such as optical flow, sift flow and divergence are also evaluated. The feature vector is computed for each of the IR videos and it is further passed on to various segmentation algorithms. In particular, the main methods used in the segmentation of fire and smoke are K-Means Clustering \cite{16}, Gaussian Mixture Models (GMM) \cite{17}, Markov Random Fields (MRF) \cite{18} and Gaussian Markov Random Fields (GMRF) \cite{19}. Finally, the confusion matrix and accuracy are computed to analyze the efficiency of the system. \par
The novelty of this approach is the following:
\begin{itemize}
    \item It involves the fusion of intensity, divergence, and optical flow-based features to obtain the most discriminative features for segmentation. The divergence and optical flow features are chosen since they calculate the flow at a given point and the displacement of pixels from one frame to another. Hence these motion features are combined with the intensity values representing the variations in temperature to form the most significant feature vector for segmentation.
    \item The cascading of the feature extraction with an MRF framework for segmentation of fire and smoke. The main advantage of the proposed system is that it does unsupervised learning by modeling the likelihood of the data. The latent variable is already known and for this experiment, it can have 3 values corresponding to fire, smoke and background. Since these unsupervised algorithms support pre-training, it can be used real-time for testing. Though various deep learning strategies are currently used to obtain high performance; they are supervised algorithms and hence require a large amount of labeled data for training \cite{39,40}. In the case of real-time firefighting scenarios, the availability of labeled data of both fire and smoke is limited.
    \item The proposed system has low computational complexity and less trainable parameters. It also requires lesser training data to obtain high accuracy in segmentation. Multi-spectral systems can be developed using these algorithms by mixing IR and UV \cite{41} data. Further, the feature extraction also allows fusing information from different sensors and hence by using this information the fire, smoke, and background can be accurately classified.

\end{itemize}

\section{Proposed Method}
The proposed technique has two main stages, which are feature extraction and segmentation. The first stage consists of characterizing every pixel of a sequence of images by several features, the first one is the magnitude (which expresses the pixel temperature at every instant) and the second one is the optical flow. This is an estimation of the speed of the particles moving in the sequence of images. This feature has two dimensions, that can be expressed in polar or rectangular coordinates. The last used feature is the divergence of the velocity vector field of the images.\par 
We use the feature images to segment the pixels into different classes in a non-supervised way. Clustering methods construct a set of class conditional likelihood functions for each of the possible classes, and posterior probabilities of the classes given the observed pixels. The segmentation is completed by taking a decision on the class of the pixel based on the maximum value for those posteriors (maximum a posteriori criterion). The possible algorithms for segmentation are K-means, Gaussian Mixture Model (GMM), Markov Random Field (MRF) and Gaussian Markov Random Field (GMRF).\par
These algorithms are based on a probabilistic model of the observable data $x_i$ as a function of its class. The data can be representative of one of $K$ classes, thus each pattern $x_i$ has an associate latent variable $z_i \in \{1, \cdots K\}$. The first two algorithms assume that the latent variables are independent and that the observable data are conditionally independent, this is $\forall i,j,~ p(x_i|x_j,z_i=k)=p(x_i|z_i=k)$. The models for K-means and GMM are usually represented by Gaussian functions. In K-means, the  $K$ class conditional are identical and isotropic. Thus, the observation log likelihood is proportional to the Euclidean distance of the samples to the $K$ means of the distributions. The posteriors are simply approximated to 1 for the distribution with the closest mean, 0 for the rest. GMM assumes variable covariance matrices, which gives more flexibility to the model, and the posteriors are computed using the bayes theorem through the data likelihood and the latent variable priors. The MRF model \cite{36, 37, 38} uses a likelihood for the data identical to the GMM one, but it assumes that there is a relationship between a pixel and its neighbors, so the latent variables in the same image are remodeled using an undirected graph. The algorithm is usually updated using the  Iterated Conditional Modes (ICM) method. The GMRF also models the likelihood in a similar way and here the random variable associated with a pixel is considered to be jointly Gaussian. The methods are summarized below. 

\subsection{Feature Extraction}
\subsubsection{Optical Flow}
The proposed fire detection algorithm takes advantage of one of the visually detectable characteristics of fire, i.e. motion. Here the motion estimation of fire is done using the Horn-Schunck Optical Flow method \cite{20}. This algorithm makes use of the flow vectors of moving objects over time to detect moving regions in an image. It computes a 2-dimensional vector known as the motion vector which indicates the velocities as well as the directions of each pixel of two consecutive frames in a time sequence. The main assumption made during this approach is intensity constancy: intensity values are preserved by moving objects from frame to frame.\par
\begin{figure}[H]
 \centering
\includegraphics[scale=.4]{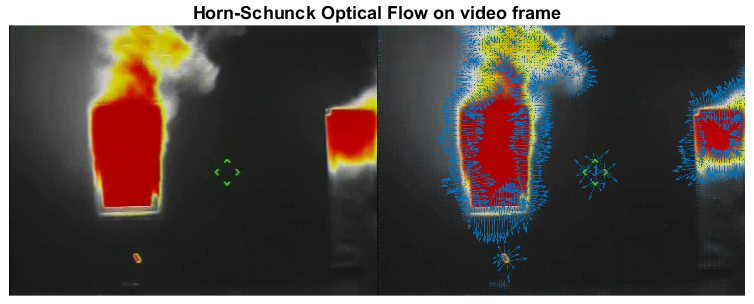}
    \caption{Optical Flow on video frame}
    \label{fig1:p1}
 \end{figure}
Another assumption in optical flow is that objects at time t will always be in the image at time t+1, but with a slight displacement. They can be expressed in terms of intensities as
\begin{equation}
     I(x,y,t)=I(x+\Delta{x},y+\Delta{y},t+\Delta{t})
\end{equation}
Using the Taylor series approximation on the above equation and by assuming the movement to be small, we obtain the following first order approximation:
\begin{equation}
\begin{split}
   I(x+\Delta{x},y+\Delta{y},t+\Delta{t})= I(x,y,t)+\\
   \frac{\partial I}{\partial x}\Delta{x}+\frac{\partial I}{\partial y}\Delta{y}+\frac{\partial I}{\partial t}\Delta{t}+\mathcal{O}(t)
\end{split}
\end{equation}
Thus the 2D motion constraint equation is given as follows:-
\begin{equation}
     I_{x}V_{x}+I_{y}V_{y}=-I_{t}
\end{equation}

\begin{equation}
\nabla I \cdot \vec{V} =-I_{t}
\end{equation}
where the velocity of the x and y components are given by $V_{x}$ and $V_{y}$ and the derivatives of the image at (x,y,t) is denoted using $I_{x}$, $I_{y}$ and $I_{t}$. The equation has no unique solution since there are 2 unknowns. This is called the aperture problem of optical flow systems.
This problem occurs when the component of the motion perpendicular to the gradient (i.e., parallel to the edge) cannot be determined. Thus Horn and Schunck added an additional constraint to perform the global regularization calculation. It was assumed that the optical flow will be smooth over relatively large areas and the objects in the image undergo only rigid motion. The regularization was done by minimizing the square of the magnitude of the gradient of the optical flow. This flow is defined as a global energy functional and later minimization is done.
\begin{equation}
     E=\iint[(I_{x}u+I_{y}v+I_{t})^2+\alpha^2(\Vert\nabla u\Vert^2+\Vert\nabla v\Vert^2)]dxdy
\end{equation}
The optical flow vector is denoted as $\vec{V}$=$[u(x,y),v(x,y)]^T$ and the regularization constant is given by $\alpha$. The parameter $\alpha$ controls the impact of the smoothing factor and thus larger values of it indicates a smoother flow. 
By solving the associated multi-dimensional Euler-Lagrange equations, the energy functional is minimized. These equations are as follows.
\begin{equation}
\begin{aligned}
\frac{\partial L}{\partial u}-\frac{\partial}{\partial x}\frac{\partial L}{\partial u_x}-\frac{\partial}{\partial y}\frac{\partial L}{\partial u_y}=0 
\\
\frac{\partial L}{\partial v}-\frac{\partial}{\partial x}\frac{\partial L}{\partial v_x}-\frac{\partial}{\partial y}\frac{\partial L}{\partial v_y}=0
\end{aligned}
\end{equation}
Here L denotes the integrand of the energy functional and further simplification gives the following two expressions.
\begin{equation}
\begin{aligned}
I_{x}(I_{x}u+I_{y}v+I_{t})-\alpha^2\nabla u=0\\
I_{y}(I_{x}u+I_{y}v+I_{t})-\alpha^2\nabla v=0
\end{aligned}
\end{equation}
The above Laplacian is approximated using finite differences and written as $\nabla u(x,y)=\bar{u}(x,y)-u(x,y)$. The weighted average of u around the pixel location (x,y) is given by $\bar{u}(x,y)$. However since the solution is dependent on the neighboring pixel values, an iterative method was devised to solve the minimization problem. The method is repeated once the neighbors have been updated and the iterative scheme derived is as follows.
\begin{equation}
\begin{aligned}
u^{k+1}=u^{-k}-\frac{I_{x}(I_{x}u^{-k}+I_{y}v^{-k}+I_{t})}{\alpha^2+I_{x}^2+I_{y}^2}\\
v^{k+1}=v^{-k}-\frac{I_{y}(I_{x}u^{-k}+I_{y}v^{-k}+I_{t})}{\alpha^2+I_{x}^2+I_{y}^2}
\end{aligned}
\end{equation}
Thus the average velocity vectors u and v are computed for each pixel in the image. Fig.~\ref{fig1:p1} shows the optical vectors on a video frame with fire and smoke.

\subsubsection{SIFT Flow}
In the optical flow algorithm, the main assumptions include brightness constancy and velocity smoothness constraint. But the pixel displacements in images of distinct scenes can be larger than the magnitude of the motion vectors. Thus, the assumptions used in classical optical flow may not be strong enough. These issues are addressed using the SIFT flow algorithm \cite{21}. Primarily the SIFT descriptors are extracted from each pixel location and these descriptors are constant with respect to the pixel displacement field. The SIFT descriptors are brightness independent and view-invariant image structures. Hence when there is significantly different image content, matching these SIFT descriptors helps to establish meaningful correspondences across the images. These descriptors can be used even when the pixel displacements are large as the image itself. But the smoothness of the pixel displacement across images is still assumed since close by pixels tend to have similar displacements. Thus, the search of the correlated SIFT descriptors across the images is formulated as an optimization problem with a cost function as follows:
\begin{equation}
\begin{split}
E(w) = {} &\sum_{p}min\Big (\Vert s_{1}(p)-s_{2}(p+w(p))\Vert_{1},t \Big )+ \\
& \sum_{p}\eta\Big (\rvert u{(p)} \lvert+\rvert v(p) \lvert \Big )+ \\
& \sum_{(p,q)\in \epsilon}min \Big (\alpha \lvert u(p)-u(q) \rvert ,d \Big )+ \\
 & min \Big (\alpha \lvert v(p)-v(q) \rvert ,d \Big)
\end{split}
\end{equation}
The above function consists of a data term, displacement term, and a smoothness term. The displacement vector at pixel location $p = (x,y)$ is given by $w(p) = (u(p),v(p))$, $\epsilon $ corresponds to the spatial neighborhood of a pixel and $ s_{i}(p) $ denotes the SIFT descriptor extracted at location p in image i. The first term in the above objective function has an L1 norm calculation to account for outliers in SIFT matching whereas a thresholded L1 norm is used in the third term along with the regularization parameter $\alpha $ to model discontinuities in the pixel displacement field. Further, the optimization is done using a dual-layer loopy belief propagation algorithm.  Here the smoothness term is decoupled and hence allows to separate u and v during message passing \cite{22}. Thus at one iteration of the message passing the complexity is reduced from $\mathcal{O}(n^4) $ to $\mathcal{O}(n^2) $. The distance transform \cite{23} is used further to reduce the complexity since the functional form of the objective function has truncated L1 norms.
\begin{figure}[t]
 \centering
\includegraphics[scale=.2]{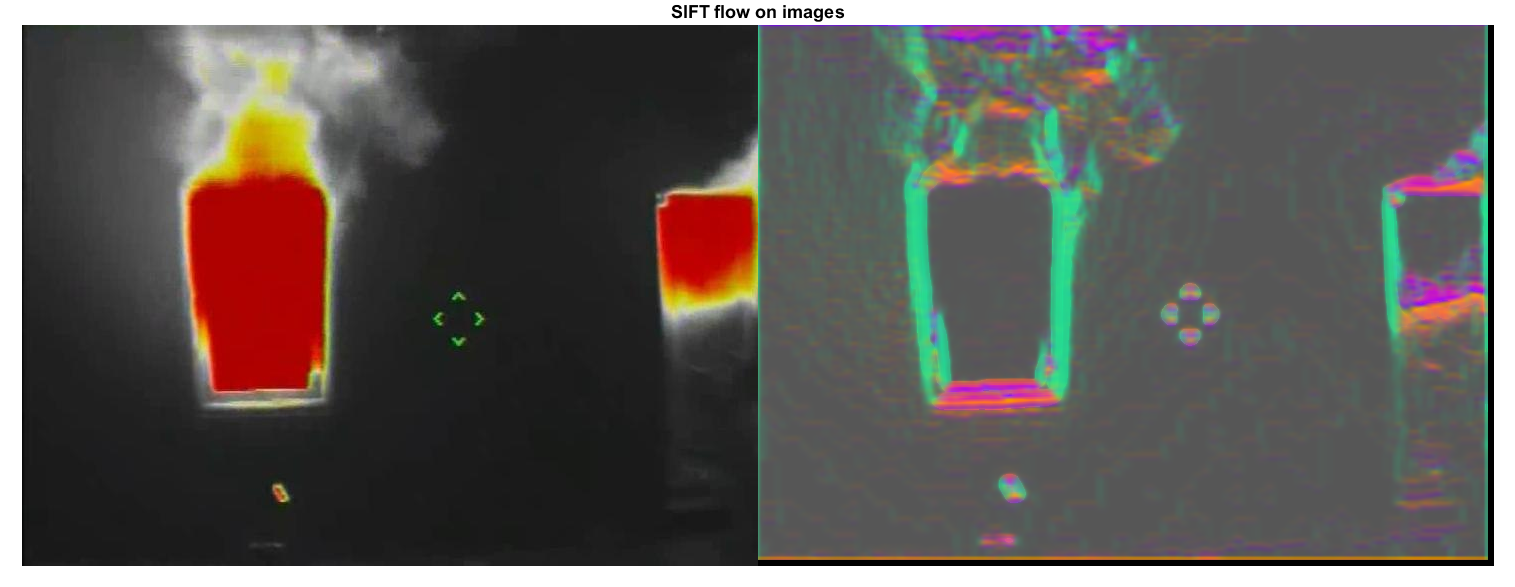}
    \caption{Sift Flow on video frame}
    \label{fig2:p2}
 \end{figure}

\subsubsection{Divergence}
The amount of flux entering or leaving a point is represented using divergence. When the flux leaves a closed surface, it is termed as positive divergence whereas flux contraction denotes negative divergence. The divergence operator inputs a vector-valued function defining a vector field and outputs the change in density of the flow at each point in the form of a scalar-valued function. Given a vector field $ \vec D=U\vec i+V\vec j $, the divergence formula is given as follows:
\begin{equation}
\begin{aligned}
div\vec D = \nabla \cdot \vec D=\frac{\partial U}{\partial x}+\frac{\partial V}{\partial y}
\end{aligned}
\end{equation}

\begin{figure}[H]
 \centering
\includegraphics[scale=.2]{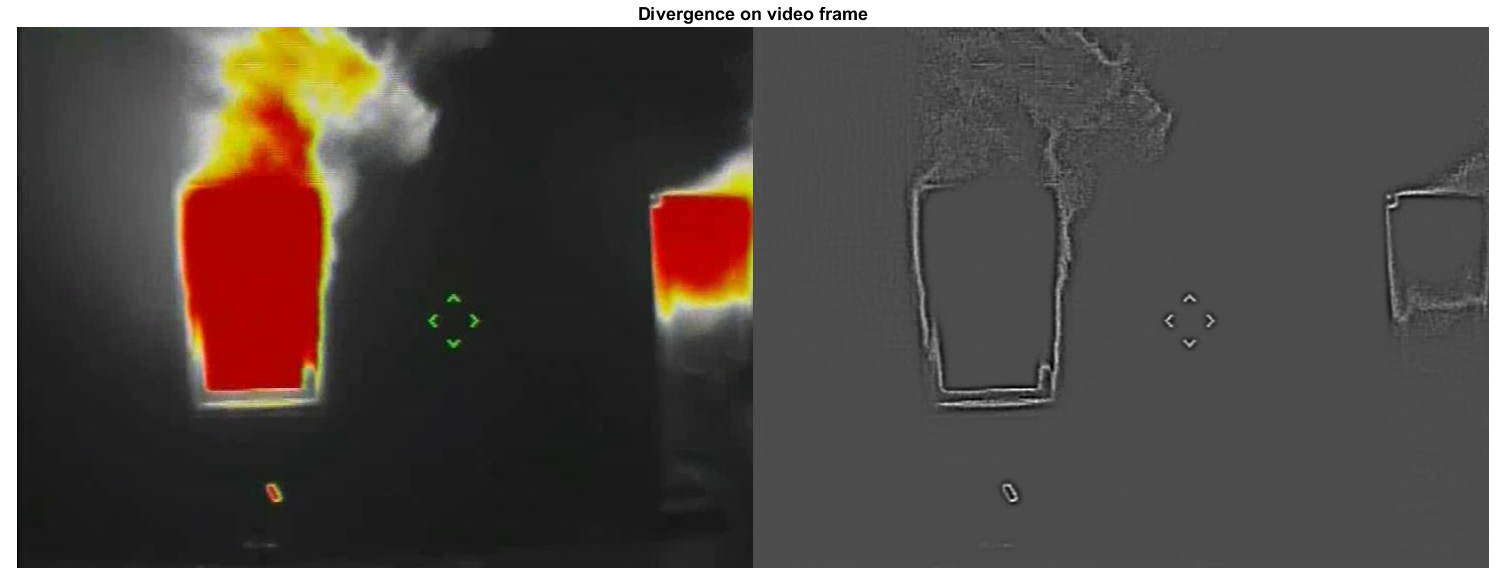}
    \caption{Divergence on video frame}
    \label{fig3:p3}
 \end{figure}

Hence the measure of expansion or compression of an object in the field is given by divergence. By applying divergence, a clear contrast between widening and narrowing of flow vectors can be visualized. The change in scale in an image is specified using divergence. Likewise, the presence of sinks and sources on the flow can also be found by applying divergence. A vector field is termed as solenoidal when it has zero divergence at every point. The values with divergence lesser than zero are termed as sinks, and greater than zero divergence is considered as the source.

\subsection{Segmentation Methods}
\subsubsection{K-means}
K-means clustering \cite{24} is an unsupervised learning algorithm. It partitions the data points into K clusters and each of the data points belongs to the cluster with the closest mean value. Based on the feature similarity, the algorithm works iteratively to assign each data point to one of K clusters. The algorithm inputs the data set and the number of clusters Κ. The initial estimates for Κ centroids are either generated randomly or selected from the data set. Then it iterates between two steps:
\begin{itemize}
\item Data assignment step:
Here each centroid defines one of the clusters. Based on the squared Euclidean distance, each data point is assigned to its nearest centroid. Let $c_{i}$ be the collection of centroids in set C, then each data point $x$ is assigned to a cluster K based on the following:
\begin{equation}
\begin{aligned}
\argmin_{c_{i}\in C}dist(c_{i},x)^2 
\end{aligned}
\end{equation}
\item Centroid update step:
The mean is calculated for all data points assigned to a centroid's cluster and thereby the centroids are recomputed.\par
\begin{equation}
\begin{aligned}
c_{i}={\frac {1}{|S_{i}|}}\sum _{x_{i}\in S_{i}}x_{i}
\end{aligned}
\end{equation} 
where $S_{i}$ be the set of data point assignments for each $i^{th}$ cluster centroid. 
\end{itemize}
The algorithm iterates between these steps and it converge when none of the data points change clusters and the sum of the distances is minimized.
\subsubsection{Gaussian Mixture Models}
A probabilistic model can be used in the representation of normally distributed subpopulation in a dataset. Gaussian mixture models \cite{17} are such models which learn about the subpopulation without knowing which subpopulation a data point belongs to. This constitutes a form of unsupervised learning since the assignment of the subpopulation is unknown. The mixture of Gaussians is represented as follows:
\begin{equation}
\begin{aligned}
p(x_{i} \lvert \theta)=\sum_{k=1}^{K} \pi_{k} \mathcal{N}(x_{i} \lvert \mu_{k},\Sigma_{k})
\end{aligned}
\end{equation}

\begin{equation}
\begin{split}
\mathcal{N}(x_{i} \lvert \mu_{k},\Sigma_{k})={} & \frac{1}{\sqrt{(2 \pi)^K\lvert \Sigma_{k}\rvert}}exp(-\frac{1}{2}(x_{i}-\mu_{k})^{T}\\
& \Sigma_{k}^{-1}(x_{i}-\mu_{k}))
\end{split}
\end{equation}
where $x_{i}$ denotes the observed variables. The mixture component weights and the component mean and covariances characterizes a Gaussian mixture model. In the multivariate case $\mu_{k} $ denotes the mean whereas $ \Sigma_{k} $ corresponds to the covariance matrix. For each latent variable $z_{k}$, we define prior probabilities $ \pi_{k} $. The total probability distribution normalizes to 1 with the constraint that $ \sum_{k=1}^{K}\pi_{k}=1 $.
\par
When the number of components K is known, expectation maximization is employed to estimate the parameters of the mixture model. It is a numerical technique used in the estimation of maximum likelihood. It is an iterative technique with the property that with each subsequent iteration the maximum likelihood of the data increases strictly. Hence it reaches a local maximum at the end of the procedure. The expectation maximization consists of two steps. In the Expectation step, the posterior probability $\gamma_{ik}$ that, each data point belongs to each cluster is calculated using the current estimated mean vectors and covariance matrices. While in the Maximization step, the cluster means and covariances are recalculated based on the probabilities calculated in the expectation step. The steps are repeated until the algorithm converges, providing a maximum likelihood estimate.
Thus, the main algorithm is as follows:
\begin{itemize}
  \item Evaluation of the log likelihood after initializing the means, covariances and the mixture component weights.
  \item E-step: Calculation of the posterior probability that the data point $ x_{i} $ belongs to component $ z_{k} $. Thus $\gamma_{ik}=p(z_{k}\lvert x_{i} ,\pi ,\mu ,\Sigma) $
\begin{equation}
\begin{aligned}
\gamma_{ik}=\frac{\pi_{k}\mathcal{N}(x_{i}\lvert \mu_{k},\Sigma_{k})}{\sum\limits_{k=1}^{K} \pi_{k} \mathcal{N}(x_{i} \lvert \mu_{k},\Sigma_{k})}
\end{aligned}
\end{equation}
 \item M-step: Re-estimate the new parameter values using the $\gamma_{ik} $ calculated in the E-step.
\begin{equation}
\begin{gathered}
\pi_{k}=\sum_{i=1}^{N}\frac{\gamma_{ik}}{N}\\
\mu_{k}=\frac{\sum\limits_{i=1}^{N}\gamma_{ik}x_{i}}{\sum\limits_{i=1}^{N}\gamma_{ik}}\\
\Sigma_{k}=\frac{\sum\limits_{i=1}^{N}\gamma_{ik}(x_{i}-\mu_{k})^2}{\sum\limits_{i=1}^{N}\gamma_{ik}}
\end{gathered}
\end{equation}
\item Evaluation of the log likelihood function using the new values of mean, covariance and mixture component weights.
\begin{equation}
\begin{aligned}
\ln{p(X\lvert \mu,\Sigma,\pi)}=\sum_{i=1}^{N}\ln(\sum_{k=1}^{K} \pi_{k} \mathcal{N}(x_{i} \lvert \mu_{k},\Sigma_{k}))
\end{aligned}
\end{equation}
\end{itemize}  
If there is no convergence,the E step is repeated and finally using the fitted model density estimation and clustering is done.

\subsubsection{Markov Random Field}
In images, neighboring pixels exhibit similar properties such as intensity, texture and color information. The Markov random field (MRF) \cite{26} is an undirected graphical model which makes use of this contextual information and represent them in probabilistic terms. Based on the Markov random field theory, any digital image consists of a discrete set of pixels which can be modeled using a set of random variables. The site is a term which is used to denote every pixel in an image and each site is given a label y which represents the intensity value of a pixel. Let an $M \times N $ digital image be described as $ S=\{(i,j)|1\leq i\leq m, 1\leq j\leq n\} $ where $S$ is a rectangular grid. The relations between the sites in S are defined using a neighborhood system and a set of sites in S is said to be a clique C if every pair of sites in C is neighbors to each other. Hence there exists two random fields; the label random field $y = \{y_{i} | s_{i} \in S\}$ and the observable random field $x= \{x_{i} |s_{i} \in S\}$. \par
The main goal of segmentation \cite{27} is to find the optimum estimation of hidden field y from observed field x i.e., to estimate the correct classification for each pixel. The MRF uses the maximum a posterior probability estimation to minimize the probability of misclassification.
\begin{equation}
\begin{aligned}
\hat{y}= \argmax_{y}P(y|x)
\end{aligned}
\end{equation}

\begin{figure*}
\centering
  \includegraphics[width=\textwidth,height=8cm]{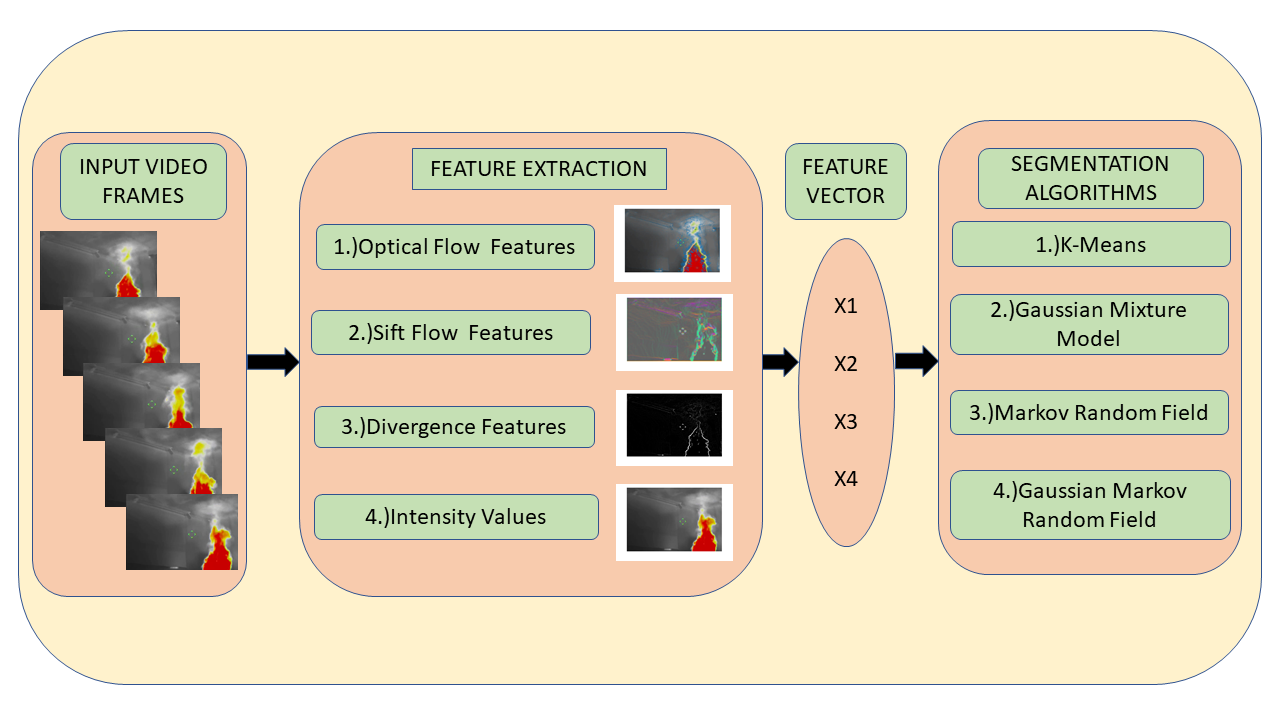}
  \caption{Structure of the Proposed Methodology}
  \label{fig4:p4}
\end{figure*}
The Hammersley-Clifford theorem \cite{25} states that any MRF can be described by a probability distribution P(y) which follows Gibbs form.
\begin{equation}
\begin{aligned}
P(y)=\frac{1}{Z}e^{-\frac{U(y)}{T}}
\end{aligned}
\end{equation}
where P(y), Z and T denotes the prior probability, normalization constant and temperature parameter respectively. The energy function U(y) can also be represented as follows.
\begin{equation}
\begin{aligned}
U(y)=\sum_{c\in C}V_{c}(y)
\end{aligned}
\end{equation}
where $V_{c}(y)$ denotes the potential function. 
Here U(y) is the sum of clique potentials $V_{c}(y)$ over all possible cliques C. It is also assumed that one pixel has at most 4 neighbors. Therefore the clique potential can be either singleton, doubleton and other higher orders depending on the number of neighbors. Thus a clique consisting of two neighboring pixels is given as follows:
\begin{equation}
\begin{aligned}
V_{c}(y_{i},y_{j})=\beta \delta(y_{i},y_{j})
\end{aligned}
\end{equation}
where $\beta$ is the coupling coefficient and when it increases the regions becomes more homogenous.\par
The segmentation problem is solved using one of MRF's pixel labeling algorithm named Iterated Conditional Modes (ICM) \cite{28}. This algorithm iteratively optimizes a statistical criterion by approximating the Maximum A-Posteriori (MAP) estimate. In the MAP approach, a posterior probability measure $P(y|x)$ and we try to find an optimal labeling $\hat{x}$ which maximizes this probability. It is also similar to minimizing the posterior energy function $U(y|x)$. ICM is thereby a greedy algorithm which tries
to find a local minimum. For each pixel, the algorithm initially provides an estimate of the labeling and it chooses the label giving the largest decrease of the energy function. The posterior energy $U(y|x)$ is given by the sum of the likelihood energy function and the prior energy function as follows:
\begin{equation}
\begin{aligned}
U(y|x)=U(x|y)+U(y)
\end{aligned}
\end{equation}

ICM, when compared with other approaches such as simulated annealing, doesn't allow the temporary increase in the potential function to obtain minimum potential. The ICM algorithm can be summarized using the following steps.
\begin{itemize}
\item Initialize by assigning an arbitrary labeling y at step n=0.
\item At step n, we find,
\begin{equation}
\begin{aligned}
y^{n+1}=\argmin_{y}U(y|x)
\end{aligned}
\end{equation}
\item Repeat the above step until convergence is obtained.
\end{itemize}

\subsubsection{Gaussian Markov Random Field}
A Gaussian Markov random field (GMRF) is an undirected gaussian graphical model with values of the random field at the nodes to be jointly Gaussian\cite{29, 35}. GMRFs fit nicely into a Bayesian framework since they are  analytically tractable. It is a continuously-valued random vector having a multivariate Gaussian distribution of the following form:
\begin{equation}
\begin{aligned}
p(y) \propto exp(-\frac{1}{2}(y-\mu)^{T}\Sigma^{-1}(y-\mu))
\end{aligned}
\end{equation}
where $\Sigma^{-1}=\Lambda$ is the the inverse covariance matrix. The quadratic form of the exponent is given as follows:
\begin{equation}
\begin{aligned}
y^{T}\Lambda y= \sum_{i} \sum_{j}y_{i}y_{j}\Lambda_{i,j}
\end{aligned}
\end{equation}
There does not exist an edge between $y_{i}$ and $y_{j}$ in the model when $\Lambda_{i,j}=0$ and hence the neighborhood system is determined by the matrix $\Lambda$. The nonzero pattern of $\Lambda$ helps to determine whether two nodes are conditionally independent. Here $\Lambda$ is sparse, that is $\Lambda_{i,j}=0$ if and only if $y_{i}$ and$y_{j}$ are conditionally independent. In practice, the GMRFs are defined using the quadratic energy function \cite{34} given by:
\begin{equation}
\begin{aligned}
U(y)=\frac{1}{2}y^{T}\Lambda y-y^{T}b
\end{aligned}
\end{equation}
where $b \in \mathbb{R}$. In the application of Bayesian image processing \cite{30}, consider the image to have a similar  $M\times N$ rectangular lattice structure as of MRFs. When a suitable prior p(y) is chosen, the maximum a posteriori (MAP) is estimated to find the optimal labels for segmentation using the ICM labeling algorithm.

\begin{figure*}{t}
\centering
\subfloat[]
  {\includegraphics[width=18cm,height=3cm]{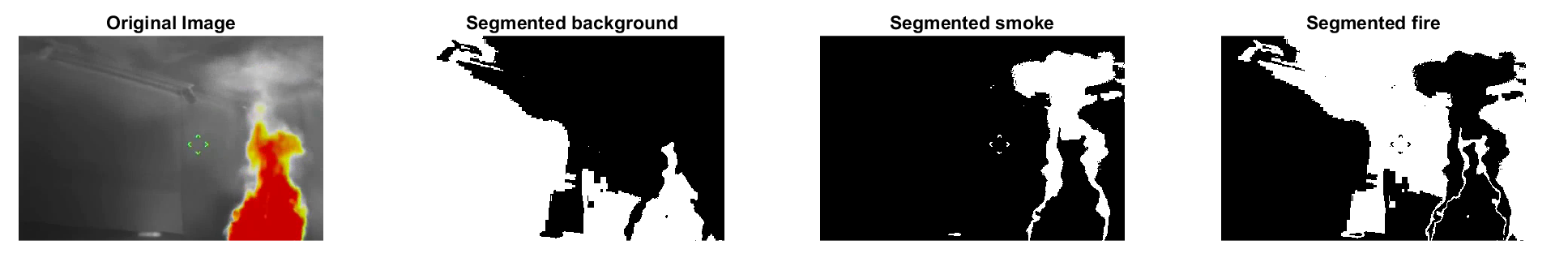}
   \label{f5:v1}}
   \hfill
\subfloat[]
  { \includegraphics[width=18cm,height=3cm]{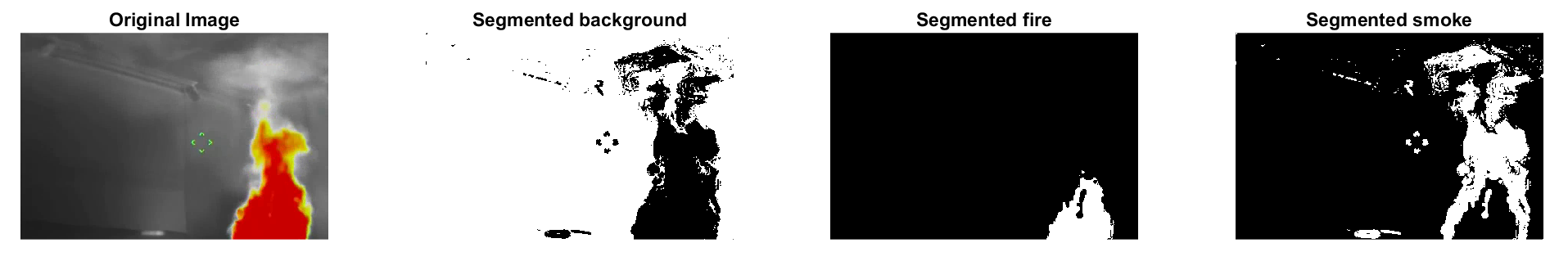}
    \label{f6:v2}}
  \hfill  
\subfloat[]
  {\includegraphics[width=18cm,height=3cm]{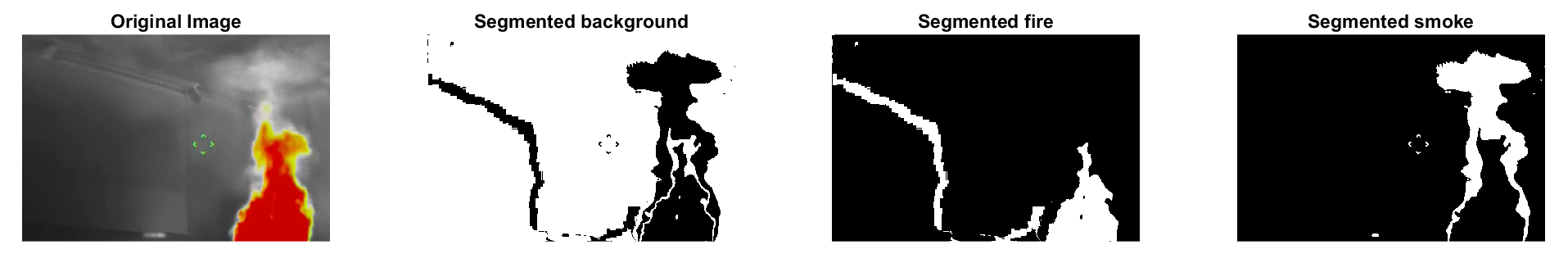}
   \label{f7:v3}}
   \hfill
\subfloat[]
  {\includegraphics[width=18cm,height=3cm]{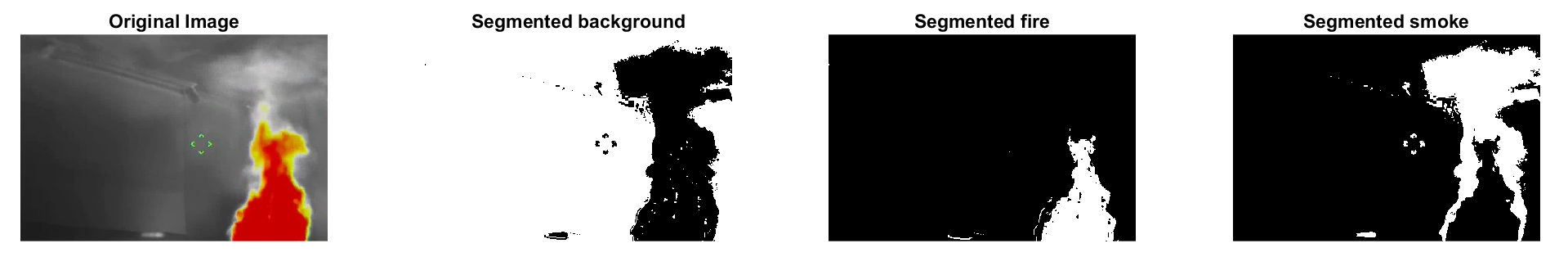}
   \label{f8:v4}}
   
\caption{Sample segmentation using (a) K-Means (b) GMM (c) GMRF (d) MRF}
\label{fig5:p5}
\end{figure*}

\section{Experiments and Results}
The proposed experimental framework reads a captured video and extracts the frame. The infra-red videos are collected from Google and the frame rate of the videos are 30fps. Here the information extracted from the first 10 frames will act as prior knowledge for the test images. Initially, the training is done using images with fire and smoke and the primary step during this phase is feature extraction. The main features used for experimentation are the intensity values of the image, magnitude of motion vectors, SIFT flow features and the divergence of the image. The intensity values are taken into consideration since fire will be having a higher intensity value compared to the smoke and background. The divergence of the image is another feature which gives the amount of flow passing through a surface surrounding a pixel. Additionally, the SIFT flow features preserve spatial discontinuities and it helps to compute pixel-wise SIFT features between two images.\par

\begin{figure*}[t]
\centering
\subfloat[]
  {\includegraphics[scale=.45]{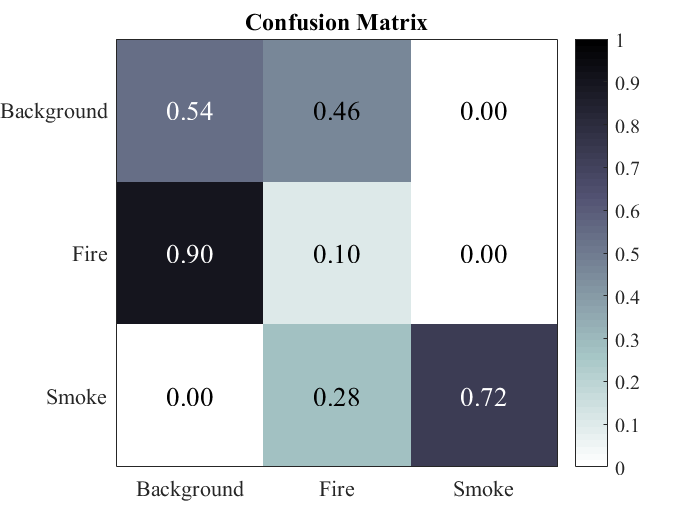}
   \label{f9:v5}}
   \hfill
\subfloat[]
  { \includegraphics[scale=.45]{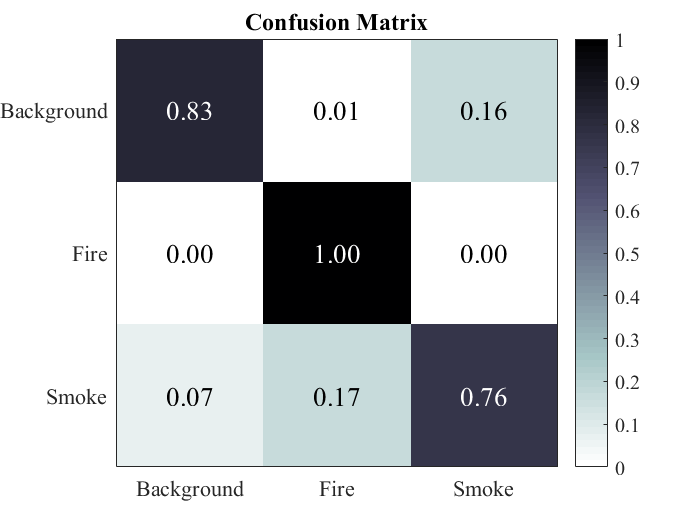}
    \label{f10:v6}}
  \vfill  
\subfloat[]
  {\includegraphics[scale=.45]{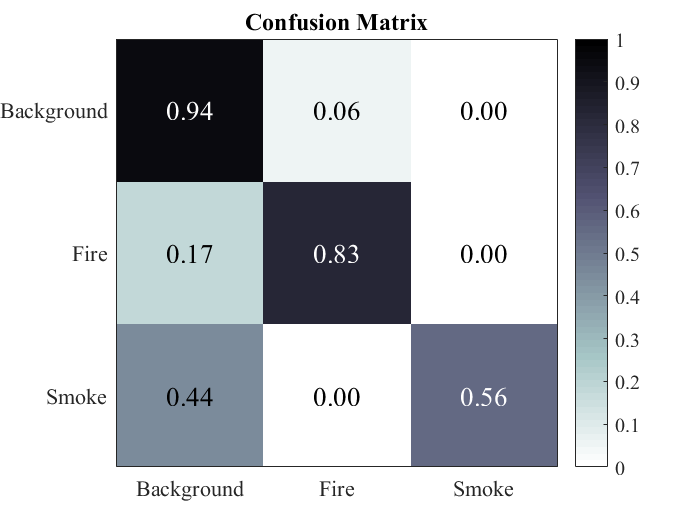}
   \label{f11:v7}}
   \hfill
\subfloat[]
  {\includegraphics[scale=.45]{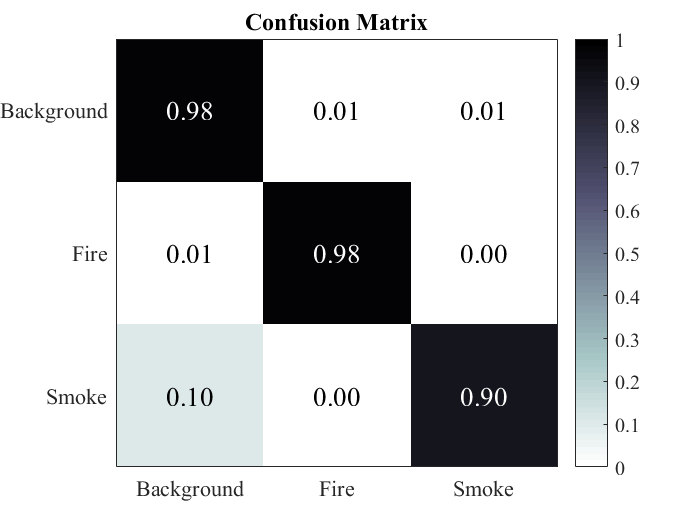}
   \label{f12:v8}}
   
\caption{Confusion matrix for (a) K-Means (b) GMM (c) GMRF (d) MRF}
\label{fig6:p6}
\end{figure*}
The motion features are computed using the Horn Schunck Optical Flow algorithm \cite{31} which computes optical flow for all the pixels in a frame. Since the optical flow is the distribution of apparent velocities of different objects in the frame. By estimating optical flow between the frames, the velocities of the objects in the video can be measured. The velocity along the x and y-direction as well as the magnitude and direction can be calculated for consecutive frames. Further various clustering techniques are used to separate the fire and smoke from the background. Thus these computed feature vectors \cite{32} are given as input to different clustering techniques such as K-Means, GMM, MRF and GMRF to segment the smoke and fire from the desired frame. Here Fig.~\ref{fig4:p4} shows the basic block diagram of the proposed algorithms. Each of this clustering \cite{33} provides the indices for the pixel values corresponding to each of the classes. These pixel indices are mapped to the original frames to perform segmentation of the region of interest. The final classification of fire, smoke, and temperature are done based on the intensity values belonging to those clusters. Hence the labeling of the clusters is done by assigning the cluster with the highest intensity values to fire, intermediate intensity values to smoke and the cluster which has a lesser intensity to background. Further, the performance evaluation is done by calculating the accuracy values for different algorithms. But the classification accuracy alone is not sufficient to select a model since it hides the details required to better understand the performance of the model. Hence the confusion matrix was computed to overcome the limitations of using only the accuracy as a decision parameter for performance evaluation. The estimation of the confusion matrix involved the manual labeling of the test frames to obtain the ground truth. Here the pixel-wise comparison is done between the segmented and ground truth values to obtain a summary of predictions made by the algorithm for each class.\par

\subsection{Sample Segmentation comparison for different algorithms}
\begin{figure*}{t}
\centering
\subfloat[]
  {\includegraphics[width=18cm,height=3cm]{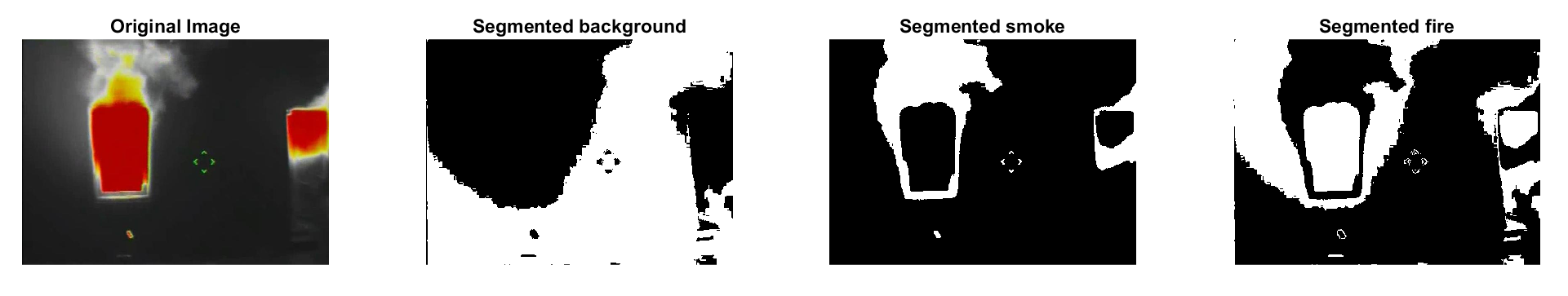}
   \label{f8:v1}}
   \hfill
\subfloat[]
  { \includegraphics[width=18cm,height=3cm]{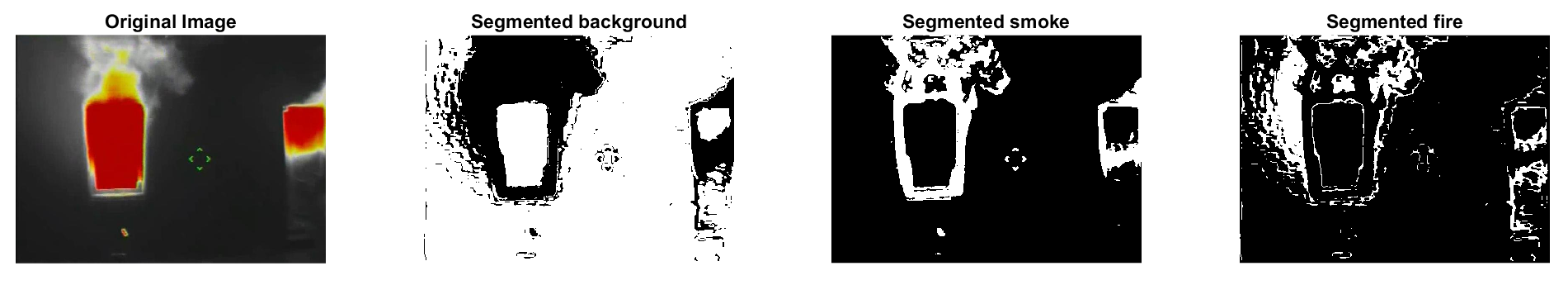}
    \label{f9:v2}}
  \hfill  
\subfloat[]
  {\includegraphics[width=18cm,height=3cm]{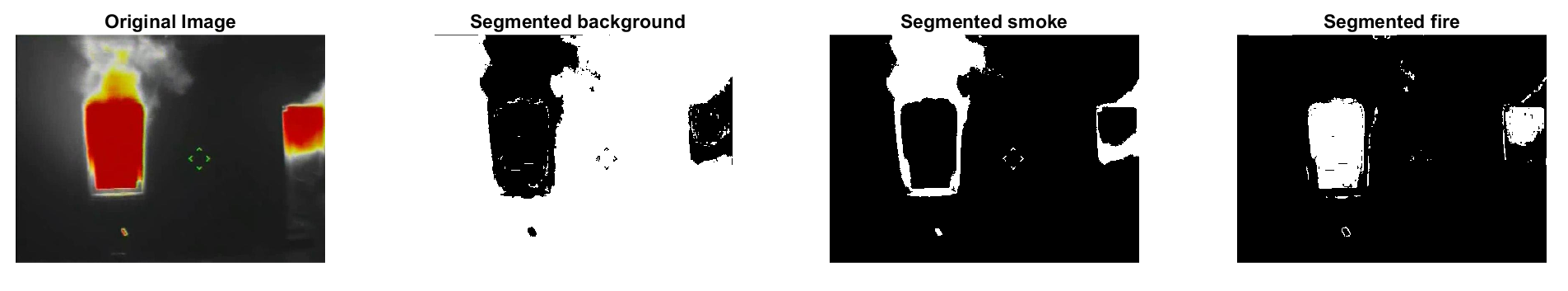}
   \label{f10:v3}}
   \hfill
\caption{Sample segmentation using (a) MRF (b) Sift Flow and MRF (c) Optical Flow, Divergence, Intensity and MRF}
\label{fig7:p7}
\end{figure*}

Fig.~\ref{fig5:p5} and Fig.~\ref{fig7:p7} shows the segmentation results using different combinations of feature extraction and clustering algorithms. The main features used for this experimentation are the divergence, intensity, sift flow and the optical flow values. It can be seen that the cascaded system using these feature vectors and MRF performs a more successful segmentation of fire and smoke than the rest of the algorithms. The rest of the segmentation algorithms misses some pixels of inner parts of smoke and fire. The MRF based approach was able to detect the smoke regions which appears blurry and indistinguishable for the human eye. Hence this prior information can be of paramount importance for the first responders during real-time fire-fighting situations.\par
In comparison with the test results shown in Fig.~\ref{fig5:p5}, it is evident that the proposed algorithm is able to disregard the unwanted artifacts from the frames. Thus, it is observed that only MRF is able to capture both the static and dynamic properties of the area of interest. The clear definition of the shape of fire and smoke will help in further analysis such as classification of fire as small, medium or large and thereby provide situational awareness.
\subsection{Confusion matrix}
\begin{figure}
\centering
  \includegraphics[scale=.45]{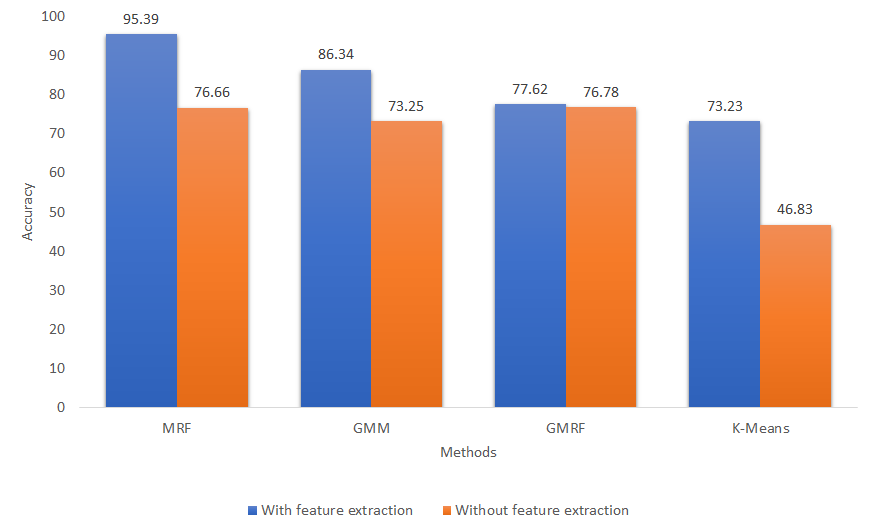}
  \caption{Accuracy of the different algorithms with and without feature extraction}
  \label{fig8:p8}
\end{figure}

The confusion matrix is an evaluation metric used widely for the analysis of semantic segmentation. It is a square matrix in which each row has instances of the true class and each column has instances of the segmented class. Hence $C_{mn}$ represents the pixels of class m which are classified as class n. Fig.~\ref{fig6:p6} shows the comparative analysis of the confusion matrix table for different methods. It can be seen that both GMM and MRF based segmentation was able to segment fire in a more accurate way than the other algorithms. In case of smoke MRF was able to perform the segmentation with 90\% accuracy whereas the rest of the methods were unable to distinguish accurately between background and smoke. Further, Fig.~\ref{fig8:p8}  shows the final calculated accuracy from the confusion matrix for the proposed methods. It can be seen that the feature extraction methods boosts the performance of the segmentation algorithms. It is also observed that the feature extraction using optical flow, divergence and intensity values and segmentation using MRF gives the highest accuracy of 95.39\%.

\section{Conclusion}
This paper introduces a novel method for fire and smoke characterization in IR images. This approach can perform unsupervised testing in real time and it can be trained parallelly in offline mode. The feature extraction methods proposed for this problem are by using optical flow, divergence, and intensity.  Even though sift flow features were tested but it did not give any significant improvement in segmentation compared to the combination of the above feature extractors. The unsupervised segmentation methods used for the comparative analysis were K-Means, GMM, GMRF, and MRF.  It was found that MRF showed better performance in the classification with a higher accuracy of 95.39\%.  It has been tested visually and quantitatively that MRF was able to distinguish fire, smoke, and background in a more precise manner. Even though GMM was able to segment most of the fire regions but it gave a much lower accuracy of 76\% for smoke segmentation. Thus, the fusion of information in the proposed method was able to produce results that outperform the traditional approaches. The future work aims to use multispectral data from UV and RGB sensors to make more accurate predictions for real-time firefighting scenarios. We would also like to extend the experimentation on dynamic and complex fire environments to test the robustness of these approaches.

\section*{Acknowledgment}
The authors would like to thank the UNM Center for Advanced Research Computing, supported in part by the National Science Foundation, for providing the high performance computing, large-scale storage, visualization resources used in this work. This work has been supported by NSF S\&CC EAGER grant 1637092.
\ifCLASSOPTIONcaptionsoff
  \newpage
\fi


\begin{IEEEbiography}[{\includegraphics[width=1in,height=1.25in,clip,keepaspectratio]{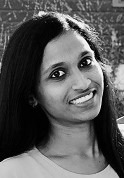}}]{Meenu Ajith} received the bachelor's degree in Electronics and Communication Engineering from Amrita school of Engineering in 2015 and the master's degree in 2017 in Electrical Engineering from The University of New Mexico in 2017. She is currently working towards her PhD degree in Electrical Engineering from The University of New Mexico. Her research interests are Machine Learning, Computer Vision, Pattern Recognition and Image Processing.
\end{IEEEbiography}

\begin{IEEEbiography}[{\includegraphics[width=1in,height=1.25in,clip,keepaspectratio]{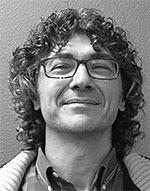}}]{Manel~Mart\'{i}nez~Ram\'on} is a professor with the ECE department of The University of New Mexico. He holds the King Felipe VI Endowed Chair of the University of New Mexico, a chair sponsored by the Household of the King of Spain. He is a Telecommunications Engineer (Universitat Politecnica de Catalunya, Spain, 1996) and PhD in Communications Technologies (Universidad Carlos III de Madrid, Spain, 1999). His research interests are in Machine Learning applications to smart antennas, neuroimage, first responders and other cyber-human systems, smart grid and others. His last work is the monographic book "Signal Processing with Kernel Methods",  Wiley, 2018.
\end{IEEEbiography}


\begin{thebibliography}{1}

\bibitem{1}B. U. Toreyin, Y. Dedeoglu, U. Gudukbay, and A. E. Cetin, "Computer vision based method for real-time and flame detection," {\em Pattern Recognition Letters}, Elsevier, 2005.

\bibitem{2}T. Chen, P. Wu, and Y. Chiou. "An early fire detection method based on image processing," {\em Proc.of IEEE International on Image Processing}, pp.
1707-1710. 2004.

\bibitem{3}Phillips, W., Shah, M., and Lobo, N.D. (2002). Flame recognition in video. {\em Pattern Recognition Letters}, 23, 319-327.

\bibitem{4}Celik, T. and Demirel, H. Fire detection in video sequences using a generic color model. {\em Fire Safety Journal}, 44(2):147-158, 2009.

\bibitem{5} Celik, T. Fast and efficient method for fire detection using image processing. {\em ETRI Journal}, 32(6), 2010. 

\bibitem{6}Wen-Bing Horng, Jian-Wen Peng and Chih-Yuan Chen, "A new image-based real-time flame detection method using color analysis," {\em Proceedings. 2005 IEEE Networking, Sensing and Control}, 2005, pp. 100-105. 

\bibitem{7}T. Celik, H. {\"O}zkaramanli and H. Demirel, "Fire and smoke detection without sensors: Image processing based approach," {\em 2007 15th European Signal Processing Conference}, Poznan, 2007, pp. 1794-1798.

\bibitem{8}Liu C.-B. and Ahuja N. Vision based fire detection. {\em In Proc. of the International Conference on Pattern Recognition (ICPR)}, volume 4, pages 134-137, 2004.

\bibitem{9}Y. Wei, Y. Chunyu, and Z. Yongming, "Based on wavelet transformation fire smoke detection method," {\em Electronic Measurement and Instruments}, 2009, vol. 2, pp. 872-875, 2009. 

\bibitem{10} D. Titterington, A. Smith, and U. Makov, {\em Statistical Analysis of Finite Mixture Distributions}. Hoboken, NJ: Wiley, 1985.

\bibitem{11} Turgay Celik, Hasan Demirel, Huseyin Ozkaramanli, and Mustafa Uyguroglu, "Fire detection using statistical color model in video sequences," {\em J. Vis. Comun. Image Represent.}, vol. 18, pp. 176-185, April 2007.

\bibitem{12}M. Zhao, A. G. Wu, C. Y. Du, "Contour Optical Flow Vector Analysis of Fire Smoke," {\em Journal of Tianjin University}.

\bibitem{13}D. Chetverikov, S. Fazekas, M. Haindl, "Dynamic texture as foreground and background," {\em Machine Vision and Applications}, vol. 22,no. 5, pp. 741-750, Sep. 2011. 

\bibitem{14}B. Uur T, "Computer vision based method for real-time fire and flame detection," {\em Pattern Recognition Letters}, vol. 27, no. 1, pp. 49-58,2006.

\bibitem{15}B. U. Toreyin, Y. Dedeoglu and A. E. Cetin, "Flame detection in video using hidden Markov models," {\em IEEE International Conference on Image Processing 2005}, 2005, pp. II-1230-3.

\bibitem{16}MacQueen, J. B. (1967). Some Methods for classification and Analysis of Multivariate Observations. {\em Proceedings of 5th Berkeley Symposium on Mathematical Statistics and Probability.} University of California Press. pp. 281-297.

\bibitem{17}Dempster A., Larid N., and Rubin D., "Maximum Likelihood From Incomplete Data via the EM Algorithm," {\em Journal of the Royal Statistical Society}, vol. 39, no. 1, pp. 1-38, 1977.

\bibitem{18}G.R. Cross, A.K. Jain, Markov random field texture models, {\em IEEE Trans. Pattern Anal. Mach. Intell.} 5 (1983) 25-39.

\bibitem{19} H. Rue and L. Held. Gaussian Markov Random Fields: Theory and Applications, {\em Monographs on Statistics and Applied Probability}. Chapman and Hall, London, 2005.

\bibitem{20} Horn, B.K.P., and Schunck, B.G. 1981. Determining optical flow, {\em Artificial Intelligence} 17: 185-204.

\bibitem{21}C. Liu, J. Yuen and A. Torralba, "SIFT Flow: Dense Correspondence across Scenes and Its Applications," in {\em IEEE Transactions on Pattern Analysis and Machine Intelligence}, vol. 33, no. 5, pp. 978-994, May 2011.

\bibitem{22}A. Shekhovtsov, I. Kovtun, and V. Hlavac. Efficient MRF deformation
model for non-rigid image matching. In {\em IEEE Conference on Computer
Vision and Pattern Recognition (CVPR)}, 2007.

\bibitem{23}P. F. Felzenszwalb and D. P. Huttenlocher. Efficient belief propagation
for early vision. {\em International Journal of Computer Vision (IJCV)},
70(1):41-54, 2006.

\bibitem{24} Hartigan, J. A.; Wong, M. A. (1979). "Algorithm AS 136: A K-Means Clustering Algorithm". {\em Journal of the Royal Statistical Society}. Series C (Applied Statistics). 28 (1): 100-108. 

\bibitem{25}S.Z. Li, Markov Random Field in Image Analysis, 3rd ed., {\em Springer}, 2009. 

\bibitem{26} S. Geman and D. Geman, "Stochastic Relaxation, Gibbs Distributions, and the Bayesian Restoration of Images," in {\em IEEE Transactions on Pattern Analysis and Machine Intelligence}, vol. PAMI-6, no. 6, pp. 721-741, Nov. 1984.

\bibitem{27} Huawu Deng and D. A. Clausi, "Unsupervised image segmentation using a simple MRF model with a new implementation scheme," {\em Proceedings of the 17th International Conference on Pattern Recognition}, 2004. ICPR 2004., 2004, pp. 691-694 Vol.2.

\bibitem{28} Besag, J," On the Statistical Analysis of Dirty Pictures",{\em Journal of the Royal Statistical Society}, vol. 48, pp. 259-302, (1986).

\bibitem{29} R. Szeliski. Bayesian modeling of uncertainty in low-level vision. {\em IJCV}, 5(3):271-301, 1990.

\bibitem{30} A. Blake, P. Kohli, and C. Rother, Markov Random Fields for Vision and Image Processing, {\em The MIT Press}, 2011.

\bibitem{31} S. Rinsurongkawong, M. Ekpanyapong, and M. N. Dailey, "Fire detection for early re alarm based on optical flow video processing," in {\em 2012 9th International Conference on Electrical Engineering/Electronics, Computer, Telecommunications and Information Technology}, pp. 1-4, May 2012.

\bibitem{32} D. Wu, N. Wang and H. Yan, "Smoke detection based on multi-feature fusion," {\em 2012 5th International Congress on Image and Signal Processing}, Chongqing, 2012, pp. 220-223.

\bibitem{33} G. Yang, H.-C. Li, and C. Liu, "Unsupervised change detection of remote sensing images using superpixel segmentation and variational gaussian mixture model," in {\em 2017 9th International Workshop on the Analysis of Multi-temporal Remote Sensing Images (MultiTemp)}, pp. 1-4, June 2017.

\bibitem{34} P. Perez, "Markov random fields and images",{\em CWI Quarterly}, vol. 11, no. 4, pp. 412-437, 1998.

\bibitem{35} M. A. T. Figueiredo and J. M. N. Leitao, "Unsupervised image restoration and edge location using compound Gauss-Markov random fields and the MDL principle," {\em in IEEE Transactions on Image Processing}, vol. 6, no. 8, pp. 1089-1102, Aug. 1997.

\bibitem{36} X. Descombes, R. D. Morris, J. Zerubia and M. Berthod, "Estimation of Markov random field prior parameters using Markov chain Monte Carlo maximum likelihood," {\em in IEEE Transactions on Image Processing}, vol. 8, no. 7, pp. 954-963, July 1999.

\bibitem{37} S. S. Saquib, C. A. Bouman and K. Sauer, "ML parameter estimation for Markov random fields with applications to Bayesian tomography," {\em in IEEE Transactions on Image Processing}, vol. 7, no. 7, pp. 1029-1044, July 1998.

\bibitem{38} H. Cao and V. Govindaraju, "Preprocessing of Low-Quality Handwritten Documents Using Markov Random Fields," {\em in IEEE Transactions on Pattern Analysis and Machine Intelligence}, vol. 31, no. 7, pp. 1184-1194, July 2009.

\bibitem{39}Y. Liu, W. Qin, K. Liu, F. Zhang and Z. Xiao, "A Dual Convolution Network Using Dark Channel Prior for Image Smoke Classification," in {\em IEEE Access}, vol. 7, pp. 60697-60706, 2019.

\bibitem{40}Z. Yin, B. Wan, F. Yuan, X. Xia and J. Shi, "A Deep Normalization and Convolutional Neural Network for Image Smoke Detection," in {\em IEEE Access}, vol. 5, pp. 18429-18438, 2017.

\bibitem{41} M. H. Tehrani, M. A. Garratt and S. G. Anavatti, "Low-altitude horizon-based aircraft attitude estimation using UV-filtered panoramic images and optic flow," {\em in IEEE Transactions on Aerospace and Electronic Systems}, vol. 52, no. 5, pp. 2362-2375, October 2016.
\end{thebibliography}
\end{document}